# Query Expansion: Term Selection using the EWC Semantic Relatedness Measure

Vitaly Klyuev
University of Aizu
Aizu-Wakamatsu
Fukushima-ken 965-8580, Japan
Email: vkluev@u-aizu.ac.jp

Yannis Haralambous
Institut Télécom – Télécom Bretagne
Dép. Informatique, UMR CNRS 3192 Lab-STICC
Technopôle Brest Iroise, CS 83818
29238 Brest Cedex 3, France
Email: yannis.haralambous@telecom-bretagne.eu

*Abstract*—This paper investigates the efficiency of the EWC semantic relatedness measure in an ad-hoc retrieval task. This measure combines the Wikipedia-based Explicit Semantic Analysis measure, the WordNet path measure and the mixed collocation index. In the experiments, the open source search engine Terrier was utilised as a tool to index and retrieve data. The proposed technique was tested on the NTCIR data collection. The experiments demonstrated promising results.

## I. Introduction

A bag-of-words representation of documents by information retrieval systems results in queries expressed utilising the language of keywords. Users face a vocabulary problem: A keyword language is not adequate to describe the information needs. Statistical analysis of user behaviors showed that the queries submitted to search engines are short (2 to 3 terms, on average) and ambiguous, and users rarely look beyond the first 10 to 20 links retrieved [20].

To improve user queries, search engines provide many tools.

Manuals and instructions are among them. They help a little, although ordinary users do not like reading.

A query suggestion feature is common for general purpose search engines. Its disadvantage is in the way to generate recommended terms. They are based on the first term typed by the user, which is is always the first one in all expanded queries [22].

The relevance feedback feature is another instrument to help users. There are at least two steps in the interaction with a search engine: The first one is the submission of the original query, and the second one is the user reaction on the results retrieved in order to provide the system with the user opinion (marking some documents as relevant). This feature is not popular among users because the mechanisms of changing queries are not clear and users cannot control the process [9].

In addition, query suggestion and relevance feedback are used to modify the queries in order to make them more accurate to express the user information needs.

This work was not supported by any organization

Query expansion is a well-known and popular technique to reformulate the user query in order to reduce the number of non-relevant pages retrieved by information retrieval systems. Another goal of query expansion is to provide the user with additional relevant documents. Automatic query expansion is an important area of information retrieval: Many scientists are involved in designing new methods, techniques, and approaches.

This paper presents authors' technique to automatically expand the user queries. This technique is based on the EWC semantic relatedness measure [21]. This measure takes into account encyclopedic, ontological, and collocational knowledge about terms. The environment for the experiments includes Terrier as a search engine and NTCIR-1 CLIR data collection for the Japanese-English cross-lingual retrieval task.

The rest of the paper is organised as follows. The next section reviews the approaches to automatic query expansion. Section 3 describe the nature of the measure used. Section 4 provides the necessary details related to this technique to expand the queries. The tools and data utilised in the experiments are presented in Section 5. The results of the experiments are discussed in Section 6 and comments on ongoing experiments are presented in Section 7. Concluding remarks are presented in Section 8.

## II. Related Work

A comprehensive review of the classical approaches to expand queries can be found in [9]. They propose different ways to obtain semantically (topically) related terms, techniques to evaluate importance of the terms found, mechanisms to define the number of terms to add (expand) the user query, and strategies to evaluate the quality of obtained results.

Generally, the semantics of the terms are clear when they are in the sentences because the meanings of the words are fixed and only one is usually selected from the set of all possible variants. However, in the queries, the terms are separate instances, and their semantics are unclear. This is called the polysemy problem. On the other hand, the same things can be described by using different terms. This is the nature of the synonymy problem. The query should be rich enough to include the possible candidates for expressing user information needs.

The general goal of query expansion is to find a solution for these two aforementioned problems. The classical solution for the synonymy problem is to apply thesauri as instruments to obtain the candidates for expansion. WordNet is widely used for this purpose [19]. Modern techniques suggest Wikipedia as a valuable source to find synonyms [12].

Many techniques are used to solve the polysemy problem. Approaches described in [13] and [16] are based on the analysis of the query log files of search engines and clicked URLs. Authors of this study [18] utilised WordNet for a deep analysis of the queries submitted to the information retrieval system in order to find the concepts and then obtain the candidate terms for expansion. The involvement of users is the feature of the approach discussed in [17]. They should select the correct ontology for each query submitted to expand the query. The authors of this study [14] also pointed out that the information exploited by different approaches differs, and combining the different query expansion approaches is more efficient than the use of any of them separately. They investigated techniques to rank the terms extracted from the retrieved documents. One is based on the measures of occurrence of the candidate and query terms in the retrieved documents. The other one utilises the differences between the probability distribution of terms in the collection and in the top ranked documents retrieved by the system. A similar idea is discussed in [15].

The authors of [10] combined the concept-based retrieval, based on explicit semantic analysis (ESA), with keyword-based retrieval. At the first step, they use keyword-based retrieval to obtain the candidates for query expansion. Then, they tune queries applying ESA. After that, they perform the final retrieval in the space of concepts.

It is difficult to compare the aforementioned approaches, because different data sets were used to evaluate them. In many cases, it is not clear wherever the test queries cover a wide range of data set topics. The performance evaluation is done automatically for some approaches, whereas for others, the authors involve the users to judge the quality of retrieval.

### III. MEASURE DESCRIPTION

In study [21], the new measure of words relatedness is introduced. It combines the ESA measure $\mu_{ESA}$ [10], the ontological WordNet path measure $\mu_{WNP}$, and the collocation index $C_\xi$. This measure is called EWC (ESA plus WordNet, plus collocations) and is defined as follows:

$$\mu_{EWC}(w_1, w_2) = \mu_{ESA}(w_1, w_2) \cdot \alpha$$
$$\alpha = (1 + \lambda_\sigma(\mu_{WNP}(w_1, w_2))) \cdot \gamma$$
$$\gamma = (1 + \lambda'_\sigma(C_\xi(w_1, w_2)))$$

where $\lambda_\sigma$ weights the WordNet path measure (WNP) with respect to ESA, and $\lambda'_\sigma$ weights the mixed collocation index with respect to ESA. This index is defined as follows:

$$C_\xi = \frac{2 \cdot f(w_1 w_2)}{f(w_1) + f(w_2)} + \xi \frac{2 \cdot f(w_2 w_1)}{f(w_1) + f(w_2)}$$

where $f(w_1, w_2)$, $f(w_2, w_1)$ are the frequency of the collocations of $w_1 w_2$ and $w_2 w_1$ in the corpus, and $f(w_i)$ is the frequency of word $w_i$. The values for constants $\lambda_\sigma$, $\lambda'_\sigma$, and $\xi$ were set to 5.16, 48.7, and 0.55, respectively.

Study [21] demonstrated the superiority of this measure over ESA on the WS-353 test set.

### IV. EXPANSION METHOD

Assume that $Z$ is a pool of term-candidates for query expansion. The formulas below present the method to select terms to expand queries. $N$ is a number of original query terms, and $j$ is an index of them. Values for the WordNet component and collocation component should be above zero in order to choose related terms. Thresholds $t_2$ for EWC values and $t_1$ are parameters adjusted in the experiments. For every word $w_i \in Z$, the weight is calculated. Word $w_i$ is selected for expansion if its weight is equal to 1.

$$weight(w_i) = \begin{cases} 1, & if \sum_{j=0}^{N} \frac{score(w_i, w_j)}{N} > t_1 \\ 0, & otherwise \end{cases}$$

$$score(w_i, w_j) = \begin{cases} 1, if \mu_{WNP} > 0 ; C_\xi > 0 ; \mu_{EWC} > t_2 \\ 0, otherwise \end{cases}$$

This approach can be interpreted as follows: A term is selected from the list of term-candidates, if the similarity score between this term and the majority of original query terms is higher than a given threshold $t_1$. The term-candidate should have non-zero values for $\mu_{WNP}$ and $C_\xi$ components.

### V. TOOLS AND DATA SETS USED

The open source search engine Terrier [1] was used as a tool to index and retrieve data. It provides the different retrieval approaches. TF-IDF and Okapi's B25 schemas [6, 9] are among them. As a data set for experiments, the NTCIR CLIR data collection [2] was used. It consists of 187,000 art-icles in English. These articles are summaries of papers presented at scientific conferences hosted by Japanese academic societies. The collection covers a variety of topics such as chemistry, electrical engineering, computer science, linguistics, and library science. The size of the collection is approximately 275.5 MB. A total of 83 topics are in Japanese. A structure of the dataset and topics is similar to that of TREC [3]. A Porter Stemmer algorithm was applied to the documents and queries, and a standard stop word list provided by Terrier was also utilised. Only the title fields were considered as a source of the queries. They are relatively short, each query consists of a few keywords. The authors of the study reported in [5] experimented with Terrier applying the same conditions to the TREC data. To measure the term similarities, an experimental tool described in [21] was utilised.

## VI. Results of Experiments

The authors implemented the proposed technique to expand queries as follows.

A straightforward approach was applied to translate queries into English: Google's translation service [4] generated queries in English. This method was selected because on-line dictionaries do not work well with terms in katakana and specific terminology [7]. Katakana is one of four sets of characters used in Japanese writing. It is primarily applied for the transcription of foreign language words into Japanese.

To obtain the candidates for query expansion, a query expansion functionality offered by Terrier was adopted. It extracts the most informative terms (in this case 10) of the top-ranked documents (in this case 3) by using a particular DFR (divergence from randomness) term weighting model [8].

Table 1 provides the list of original queries (topics 1, 12, and 24), terms-candidates for expansion (arranged by the decreasing score calculated by Terrier), and the final sets of terms used to expand queries (they are in bold). One to five terms were selected by this method. As one can see from this table, this technique does not usually select the top-ranked terms as candidates for expansion from the Terrier engine point of view.

As mentioned in Section V, a total of 83 topics are available to retrieve documents from the collection. The original goal of topics 0001 to 0030 is to tune the parameters of the retrieval system. Their relevance judgments are known in advance. Topics 0031 to 0083 were used in official runs at the NTCIR 1 Workshop. Organisers found that the number of relevant documents for 13 topics of the 53 contained less than five relevant documents per topic in cross-lingual retrieval. Hence, they discarded these topics from evaluation [25]. The full set was used in these experiments because the goal is to compare the performance of different methods implemented in the same environment. In the evaluations, the partially relevant documents were considered as irrelevant.

To archive this, the corresponding file was applied when evaluating the retrieval results.

Table 2 summarises the results of retrieval to tune thresholds $t_1$ and $t_2$. The test queries were generated from topics 0001 to 0030. It is important to note that when the queries are expanded with all the terms proposed by Terrier, the retrieval results drop to zero. The retrieval utilising the TF-IDF schema produced better results for the original queries (without expansion) compared to the BM25 and InL2 models [1]. The line *system* shows this result. The first number in the cells is the value of average precision, and the second one is the value of R-precision. The performance of retrieval with expanded queries utilising the ESA and EWC approaches for the threshold values ($t_1$ equals 0.67 and $t_2$ is ranges from 0.08 to 0.15) is better compared to the variant without expansion. For the EWC measure, the maximum of the retrieval performance is reached when the values of thresholds $t_1$ and $t_2$ are set to 0.67 and 0.12. For ESA, the optimal threshold values are 0.67 and 0.08. The performance of ESA is higher than EWC.

Table 3 summarises the results of retrieval for topics 31 to 83. The threshold values were set to the optimal parameters (see Table 2). Six runs were executed. Figure 1 shows an averaged precision/recall graph across 53 queries. The EWC measure demonstrated better performance over ESA in both cases. The line *system* shows the retrieval results without expansion for TF-IDF and BM25 schemes.

To summarize, one can conclude that the EWC measure provides little benefit over ESA, as the results of the retrieval are better. Ontological knowledge combined with collocational knowledge helps in the selection of expansion terms.

## VII. Ongoing Experiments

The authors are conducting experiments on the translation of queries from Japanese into English. The Mecab system [24] is utilised to segment the queries and extract Japanese terms. An on-line dictionary Space ALC [23] was also applied to retrieve all English variants for the corresponding Japanese terms. The EWC metrics are applied to pairs in order to select the most closely related elements from the sets of the meanings (terms with the highest sum of weights between pairs). For the queries consisting of only one term, the most common English variant is selected. Queries generated in this way will be submitted to the search engine, and comparison will be done with the results of retrieval for queries obtained from Google's translation service.

## VIII. Conclusion

This study tested the semantic relatedness measure when selecting the terms to expand queries. Key components of this measure are the ESA measure, the WordNet path measure, and the mixed collocation index. Results produced by the Terrier search engine were a base line in the experiments.

TABLE I.
TOPICS: 1, 12 AND 24: ORIGINAL AND EXPANDED QUERIES

| Topic | Original query | Terms for expansion |
|---|---|---|
| 1 | Robot | **Robot** |
| | | person |
| | | **human** |
| | | multi |
| | | comput |
| | | sice |
| | | design |
| | | will |
| | | confer |
| | | paper |
| 12 | Mining methods | Mine |
| | | method |
| | | rule |
| | | data |
| | | databas |
| | | associ |
| | | discoveri |
| | | **larg** |
| | | tadashi |
| | | solv |
| | | amount |
| 24 | Machine translation system | Machin |
| | | translat |
| | | **system** |
| | | exampl |
| | | **base** |
| | | **masahiro** |
| | | **method** |
| | | nation |
| | | convert |
| | | **problem** |

TABLE II.
THRESHOLDS TUNING: TOPICS 1 TO 30

| t1 | t2 | EWC: Average Precision R-Precision | | | ESA/ BM25 |
|---|---|---|---|---|---|
| | | InL2 | TF-IDF | BM25 | |
| 0.5 | 0.1 | 0.2940 0.3216 | 0.3031 0.3314 | 0.3072 0.3324 | **0.3101** **0.3347** |
| 0.65 | 0.09 | 0.2940 0.3300 | 0.2936 **0.3332** | 0.2955 0.3278 | **0.2961** 0.3276 |
| 0.67 | 0.07 | **0.2973** **0.2977** | 0.2954 0.2963 | 0.2960 0.2916 | 0.2959 0.2916 |
| 0.67 | 0.08 | 0.3101 0.3277 | 0.3151 0.3268 | 0.3140 0.3265 | **0.3172** **0.3315** |
| 0.67 | 0.09 | 0.3073 0.3256 | 0.3105 **0.3373** | **0.3106** 0.3352 | 0.3080 0.3373 |
| 0.67 | 0.1 | 0.3030 0.3295 | **0.3103** **0.3350** | 0.3099 0.3349 | 0.3099 0.3318 |
| 0.67 | 0.11 | 0.3049 0.3239 | **0.3121** **0.3309** | 0.3110 0.3292 | 0.3102 0.3302 |
| 0.67 | 0.12 | 0.3049 0.3239 | **0.3121** **0.3309** | 0.3110 0.3292 | 0.3092 0.3284 |
| 0.67 | 0.13 | 0.3049 0.3125 | **0.3114** 0.3245 | 0.3099 0.3248 | 0.3111 **0.3282** |
| 0.67 | 0.15 | 0.3033 0.3143 | **0.3115** 0.3267 | 0.3111 **0.3282** | 0.3110 0.3282 |
| 0.69 | 0.09 | 0.3073 0.3256 | 0.3105 **0.3373** | **0.3106** 0.3352 | 0.3080 0.3373 |
| 0.75 | 0.1 | 0.3030 0.3295 | 0.3103 0.3330 | **0.3099** **0.3349** | 0.3099 0.3318 |
| System | system | 0.2980 0.2995 | 0.3034 0.3166 | 0.3017 0.3163 | |

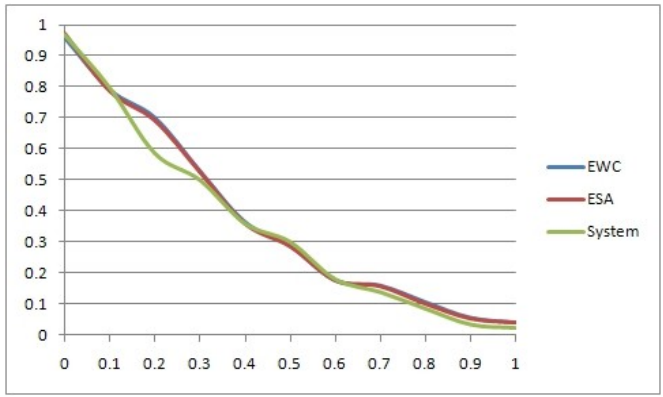

Fig 1. Averaged 11-point precision-recall graph across topics 31 to 83

TABLE III.
RETRIEVAL RESULTS: TOPICS 31 TO 83

| t1 | t2 | EWC: Average Precision R-Precision | | ESA: Average Precision R-Precision | |
|---|---|---|---|---|---|
| | | BM25 | TF-IDF | BM25 | TF-IDF |
| 0.67 | 0.08 | **0.2363** **0.2416** | | 0.2349 0.2390 | |
| 0.67 | 0.12 | | 0.2200 0.2317 | | 0.2161 0.2284 |
| System | system | 0.2225 0.2350 | **0.2218** **0.2359** | | |

Term candidates for the expansion were also generated by Terrier. The proposed techniques were applied to the ad-hoc retrieval task. As a data set, the NTCIR-1 CLIR Test collection was used. The initial English queries were obtained automatically applying Google translate. The queries were expanded by applying the Wikipedia-based Explicit Semantic Analysis measure, and the DFR mechanism, and the semantic relatedness measure. The retrieval results showed superiority of the last one over ESA and DFR.